\documentclass[letterpaper]{article} 
\usepackage[preprint]{aaai2027}  
\usepackage[hyphens]{url}  
\usepackage{graphicx} 
\usepackage{natbib}  
\usepackage{caption} 
\usepackage{booktabs}
\usepackage[table]{xcolor}
\definecolor{highlighter}{RGB}{230,230,230}

\usepackage{etoolbox}
\makeatletter
\patchcmd{\@maketitle}{These authors contributed equally.}{Equal contribution\hspace{1.5em}\textsuperscript{\textdagger} Corresponding author\newline\hspace*{2.1em}Preprint.}{}{\errmessage{Cannot patch aaai2027 equalcontrib text (1)}}
\patchcmd{\@maketitle}{These authors contributed equally.}{Equal contribution\hspace{1.5em}\textsuperscript{\textdagger} Corresponding author\newline\hspace*{2.1em}Preprint.}{}{\errmessage{Cannot patch aaai2027 equalcontrib text (2)}}
\makeatother

\usepackage{amsmath}
\usepackage{amssymb}

\usepackage{array}
\usepackage{tabularx}
\newcolumntype{C}{>{\centering\arraybackslash}X}

\title{Do Dynamic Routers Need Memory?\\HeRo: History-Aware Routing for Efficient LLM Inference}
\author{
    Hongjin Lin$^{1}$\equalcontrib, Wentao Wan$^{1}$\equalcontrib, Keze Wang$^{1}$\textsuperscript{\textdagger}
}
\affiliations{
    $^1$ Sun Yat-sen University\\
    linhj53@mail2.sysu.edu.cn, \textbraceleft wanwentao93,kezewang\textbraceright@gmail.com
}

\begin{document}

\maketitle

\begin{abstract}
 Dynamic layer routing reduces the inference cost of Large Language Models (LLMs) by learning to skip layers for individual tokens. Existing methods, however, treat each routing decision as a local operation conditioned solely on the current hidden state which is a formulation that overlooks the sequential, path-dependent nature of routing across depth: earlier decisions shape the representations seen by downstream routers, and the layer-usage objective couples all decisions jointly. We propose \textbf{H}istory-Awar\textbf{e} \textbf{Ro}uting (\textbf{HeRo}), a dynamic routing framework that resolves this mismatch by introducing a \textbf{router memory mechanism} to maintain an explicit routing state across model depth. The memory is constructed via \textbf{linear attention}, incrementally aggregating preceding routing scores and their induced residual updates into a compact history representation. At each routed layer, the router conditions jointly on this accumulated state and the current hidden representation to select the executed branch. Instantiated for token-wise FFN routing, HeRo trains only lightweight routers and adapters on a \textbf{frozen backbone}, requiring no modification to pretrained parameters. Across Llama 3.1-8B, Llama 2-7B, and Llama 2-13B, HeRo consistently achieves the \textbf{highest aggregate performance retention among ten baselines}. On Llama 3.1-8B, \textbf{it bypasses 26.87\% of model parameters while achieving 100.24\% of dense model performance} across seven benchmarks, and \textbf{retains 97.01\% while bypassing 38.82\%} of model parameters under a tighter computation budget. Ablation studies confirm that removing routing history consistently degrades performance, most notably on multistep reasoning and code generation, validating that \textbf{explicit routing memory} enables more accurate and adaptive dynamic routing than solely conditioning on hidden state.
\end{abstract}

\section{Introduction}

 Large Language Models (LLMs) generate text autoregressively by propagating hidden state through the full Transformer decoder stack and predicting the next token \citep{vaswani2017attention}. This static-depth regime is computationally inefficient because the full decoder stack is applied uniformly to every next-token prediction, even though different predictions require different amounts of computation \citep{men-etal-2025-shortgpt,kim2024shortened}. Empirically, predictions that are highly predictable from the available context can often be generated with fewer layers, whereas those that depend on multistep reasoning or information distributed across long contexts require more layers of computation \citep{schuster2022calm,raposo2024mixture}.

\begin{figure}[t]
\centering
\includegraphics[width=\columnwidth]{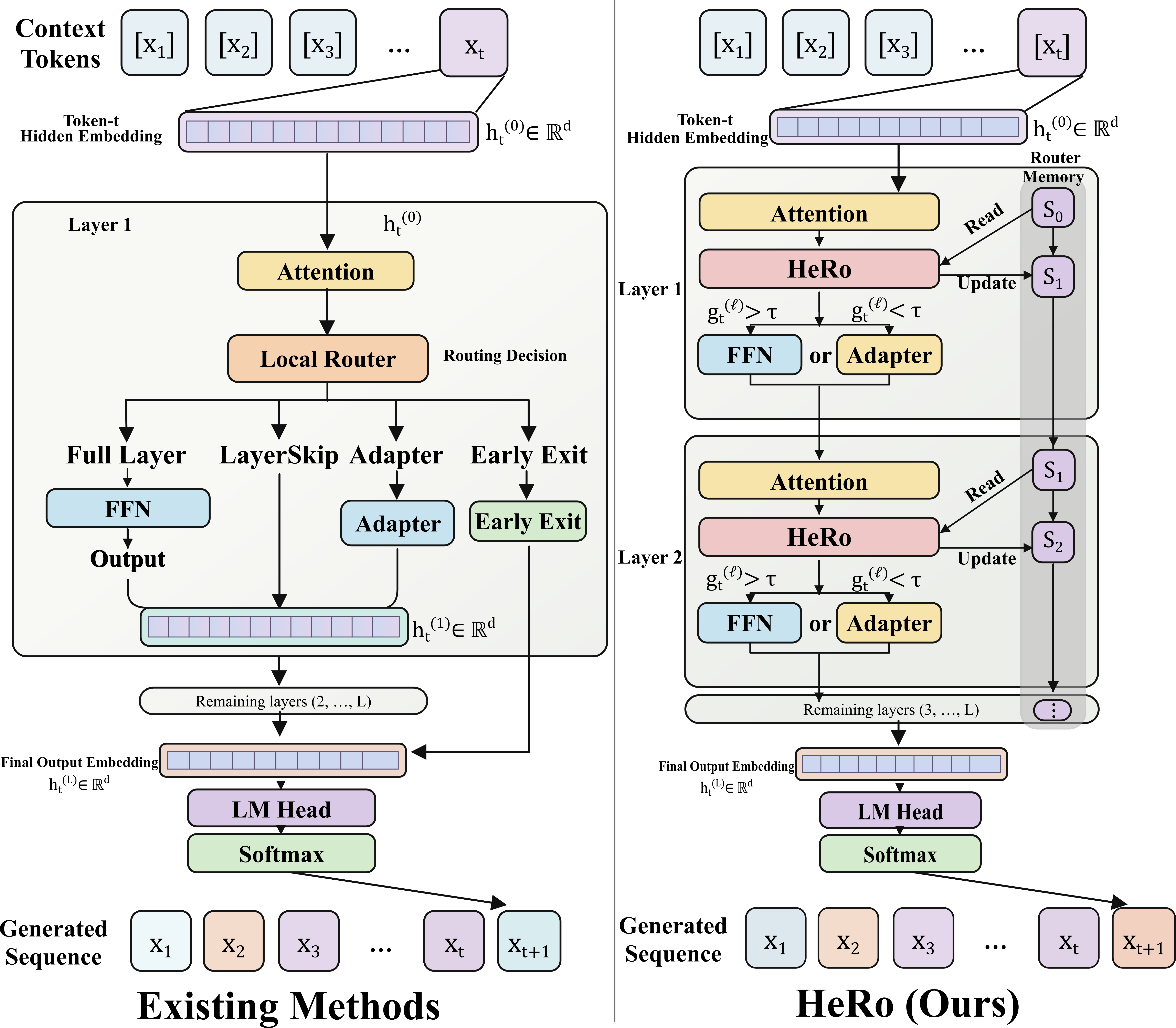}
\caption{Dynamic feed forward network (FFN) routing proceeds as a sequence of decisions along model depth.
HeRo augments the current token representation with a recurrent state constructed from preceding routing scores and changes in the residual stream, and uses this state to select the full FFN or adapter branch.}
\label{fig:intro}
\end{figure}

 Prior work seeks to reduce computation across model depth through three broad strategies: static pruning, early exit, and dynamic routing. Static pruning removes redundant layers or parameters before inference and produces a fixed smaller model while still applying the same computation to every input \citep{men-etal-2025-shortgpt,kim2024shortened,ma2023llmpruner,ashkboos2024slicegpt}. Early exit methods select an intermediate exit layer according to confidence estimated from the hidden state at that layer, allowing the executed depth to vary across predictions while restricting computation to a contiguous prefix of the decoder \citep{schuster2022calm,bae-etal-2023-free,elhoushi-etal-2024-layerskip}. Dynamic routing, including dynamic layer skipping and Mixture of Depths, uses learned routers or gates to decide which modules are executed and can bypass selected intermediate layers or modules while continuing through subsequent layers, thereby reducing computation without discarding all later transformations \citep{raposo2024mixture,he-etal-2025-router,luo2025flexidepth,heakl2026drllm,zhao2025skipgpt,laitenberger2026what}.

As illustrated in the left part of Figure~\ref{fig:intro}, existing dynamic routing methods typically formulate the routing decision at layer $l$ as a local decision conditioned solely on the current representation $\smash{h_t^{(l)}}$ of token $t$. Yet the training objective regularizes routing scores across depth, jointly optimizing each layer's decision. For each token, routing decisions at preceding layers determine which residual updates are applied and thereby shape the hidden states presented to subsequent routers. Moreover, the layer-usage loss jointly regularizes the routing scores across all routed layers, so the decision at layer $l$ depends on both the preceding routing history and the remaining routing horizon. Dynamic routing therefore constitutes a sequential decision process along the model depth, in which each routing decision is conditioned on the preceding routing history, which comprises both the routing scores assigned at preceding layers and the residual updates induced by those decisions.

\begin{figure*}[t]
\centering
\includegraphics[width=0.88\textwidth]{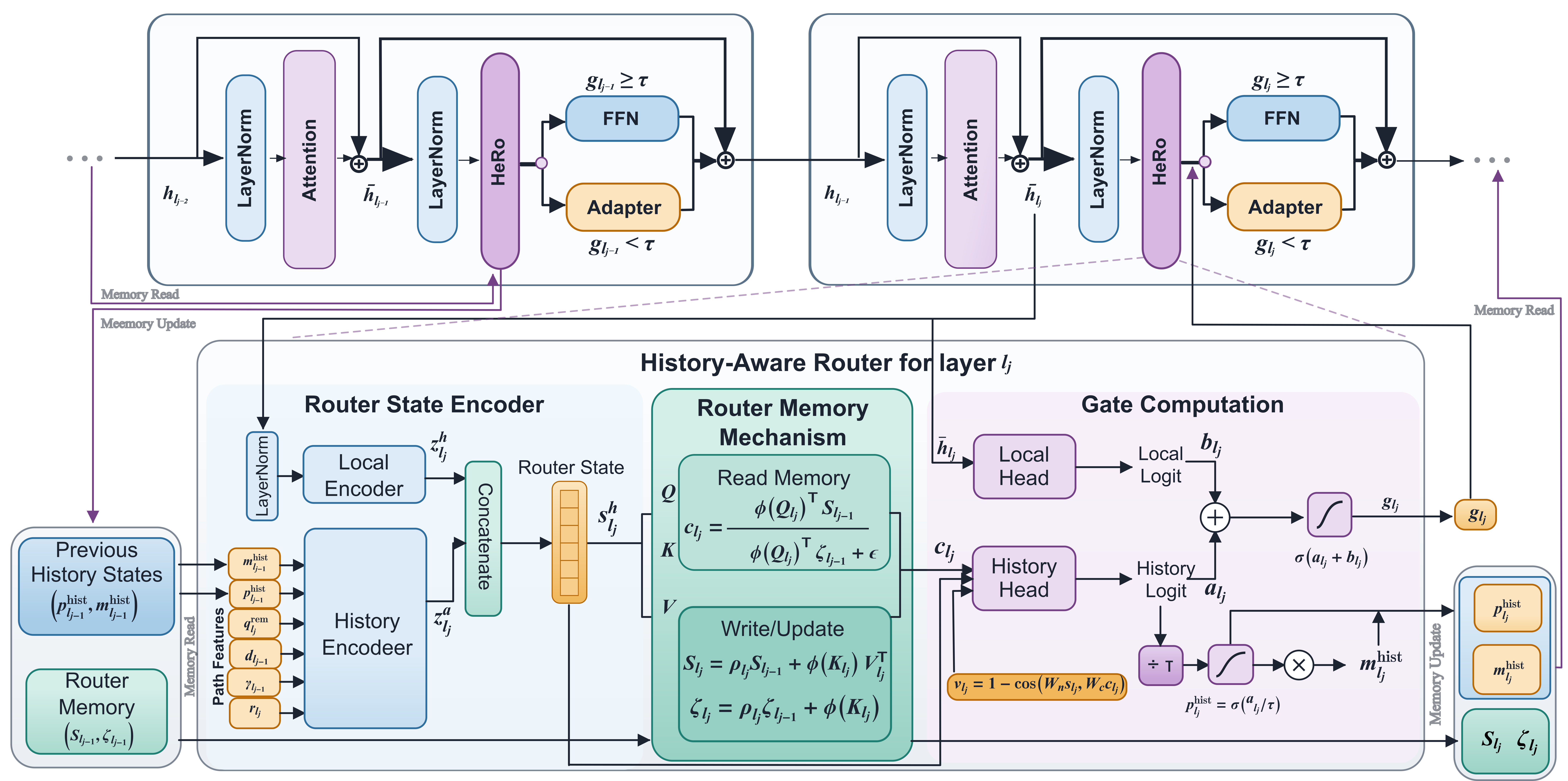}
\caption{Overview of HeRo across two routed blocks. At each block, the router encodes the current post attention representation, remaining routed depth, preceding routing scores, and completed residual stream transitions. It reads the router memory, combines the resulting history logit with a local logit, and thresholds the gate to select the pretrained FFN or adapter.}
\label{fig:overview}
\end{figure*}

Although $\smash{h_t^{(l)}}$ reflects the cumulative effects of preceding routing decisions, it does not explicitly record the preceding routing history. A router conditioned only on $\smash{h_t^{(l)}}$ must therefore infer that history from its cumulative effect on the current hidden state. Residual-stream analyses show that latent belief-state structure can be encoded in hidden representations \citep{shai2024controlled}, but this does not establish that a lightweight routing head can recover the explicit sequence of earlier gate choices. We therefore treat $\smash{h_t^{(l)}}$ as an incomplete \emph{observed} state for routing: it omits the directly observable gate scores and residual transitions available along the executed path. Studies of Mixture-of-Experts routing further show that propagating a routing state across layers provides information for expert selection that is complementary to the current hidden state alone \citep{qiu2025rmoe}. These observations motivate maintaining an explicit routing state across depth that summarizes the preceding routing history and the residual updates induced by routing decisions.

We propose History-Aware Routing (HeRo), a dynamic routing framework that incorporates a router memory mechanism based on linear attention over model depth to maintain an explicit state of preceding routing decisions and their effects on the residual stream. For each token, this memory incrementally aggregates layerwise routing states produced by preceding routed layers. At routed layer $l$, the router predicts a routing score from the current hidden representation $\smash{h_t^{(l)}}$ together with the memory state, and compares it with a threshold to make the routing decision. The routing score and the induced residual update from the selected branch are then encoded into a new layerwise routing state and written back into the memory. We instantiate HeRo for FFN routing, selecting between the pretrained FFN and a lightweight adapter at each routed layer while keeping all attention sublayers dense.

Our contributions are as follows:
\begin{itemize}
\setlength{\itemsep}{0.15em}
\setlength{\parsep}{0pt}
\setlength{\parskip}{0pt}
\setlength{\topsep}{0.2em}
 \item We characterize dynamic routing in Transformers with decoder-only architectures as a \textbf{path dependent sequential decision process} across model depth. Each routing decision determines which residual update is applied and therefore changes the representation presented to downstream routers, while the routing objective couples decisions across routed layers. This characterization exposes a mismatch in existing routers: routing decisions are interdependent across depth, yet each router is conditioned only on the current hidden representation, which does not explicitly preserve the routing history that shaped it.
\item We propose History-Aware Routing (HeRo), a dynamic routing framework that addresses this mismatch by introducing a \textbf{router memory mechanism} that maintains an explicit state of preceding routing decisions and their effects on the residual stream. HeRo uses \textbf{linear attention} over model depth to incrementally aggregate layerwise routing states formed from prior routing scores and the residual updates produced by the selected branches. This memory is updated at each routed layer and provides a compact representation of routing history. Each router conditions its decision jointly on the current hidden representation and the accumulated routing state. We instantiate HeRo for FFN routing while retaining dense computation in all attention sublayers.
\item Across three model backbones, HeRo achieves the \textbf{highest performance retention} among the evaluated parameter skipping methods. On Llama 3.1-8B, \textbf{it achieves 100.24\% of dense model performance while bypassing 26.87\% of parameters} and \textbf{retains 97.01\% while bypassing 38.82\%} under the tighter computation budget. Ablations further show that routing history improves performance across most benchmarks, with particularly notable gains on multistep reasoning tasks. These results show that the router memory mechanism enables more accurate and selective routing by incorporating information accumulated across preceding layers into each routing decision.
\end{itemize}

\section{Related Work}

\noindent\textbf{Adaptive Depth for Efficient Language Modeling.}
Transformer depth can be reduced before inference through static compression or during inference through input dependent computation. Static pruning identifies redundant layers or structural components and uses the same retained structure for every input \citep{men-etal-2025-shortgpt,kim2024shortened,ma2023llmpruner,ashkboos2024slicegpt}. Adaptive methods retain the full stack and determine how much of it to execute for each input or token. Early exit terminates computation when an intermediate prediction satisfies a confidence criterion. Developed initially for encoder models, this strategy was later adapted to autoregressive generation by CALM and FREE \citep{xin-etal-2020-deebert,liu-etal-2020-fastbert,zhou2020bert,schuster2022calm,bae-etal-2023-free}. LayerSkip combines layer dropout with an early exit objective to support speculative decoding with the same model \citep{elhoushi-etal-2024-layerskip}. Layer skipping uses input-dependent routing to bypass intermediate blocks while continuing computation in later blocks \citep{liu2024unified,raposo2024mixture,luo2025flexidepth}. This progression shifts the focus from how much depth to remove globally to which computations each input should receive.

\noindent\textbf{Dynamic Layer and Module Routing.}
Layer routing spans token, prompt, and sequence level decisions. Mixture of Depths assigns tokens to blocks under capacity constraints, while FiRST, FlexiDepth, and Dr.LLM route at prompt, token, and sequence scope, respectively \citep{raposo2024mixture,jain2024first,luo2025flexidepth,heakl2026drllm}. Module routing targets attention and MLP computation. FFN SkipLLM and DiffSkip bypass MLPs while retaining attention, whereas SkipGPT learns separate policies for attention and MLP components \citep{jaiswal2024ffnskipllm,luo-etal-2025-diffskip,zhao2025skipgpt}. Router Tuning trains only routers for attention and Mixture of Experts modules, while residual gating converts learned module scores into hard skips \citep{he-etal-2025-router,laitenberger2026what}. HeRo routes between the pretrained MLP and a lightweight adapter while keeping attention active.

\noindent\textbf{Routing with Cross Layer State.}
\citet{glavas2024dynamic} find that decoder hidden states do not consistently outperform learned static vectors for token level layer selection. RMoE propagates a recurrent state across consecutive Mixture of Experts routers to model dependencies between expert assignments \citep{qiu2025rmoe}. LIMe, MRLA, and Vertical Attention aggregate representations across depth within the backbone \citep{gerasimov2025lime,fang2023mrla,kojima2025vertical}. HeRo uses prior gate scores and residual transitions to construct a cross layer routing state.

\section{Method}

Following the sequential decision formulation introduced in the Introduction, HeRo processes each routed block through four operations, as illustrated in Figure~\ref{fig:overview}. At the start of a forward pass, the depth memory is initialized to zero. Upon reaching routed block $l_j$, the router first encodes the post-attention representation $\bar{h}_{l_j}$ alongside scalar path features (capturing routing progress, the latest completed residual transition, and prior history scores) into a unified router state $s_{l_j}$. Using a query derived from this state, it then reads the accumulated memory to retrieve a history context $c_{l_j}$ summarizing states from preceding routed blocks. A history head consumes the state and context, while a local head consumes $\bar{h}_{l_j}$, producing logits that are summed and thresholded to select the frozen FFN or lightweight adapter. Finally, the router writes the state back into memory via the forgetting recurrence; the current decision and its residual effect enter the path features at the next routed block. The following subsections formalize these operations.

\subsection{Router Computation}

We consider a Transformer decoder with $L$ blocks and apply FFN routing
at $L_R$ selected blocks, indexed in depth order by
$1\leq l_1<\cdots<l_{L_R}\leq L$. All other blocks retain their
pretrained computation. Attention remains dense at every routed block;
the representation after attention and the normalized FFN input are
\begin{equation}
\begin{aligned}
\bar{h}_{l_j}
&=h_{l_j-1}+\mathrm{Attn}_{l_j}\!\left(
\mathrm{LN}^{\mathrm{attn}}_{l_j}(h_{l_j-1})\right),\\
u_{l_j}
&=\mathrm{LN}^{\mathrm{ffn}}_{l_j}(\bar{h}_{l_j})
\end{aligned}
\label{eq:dense_attention}
\end{equation}
At each routed block, the router selects either the frozen pretrained
feedforward network (FFN) or a trainable bottleneck adapter. The adapter
uses down and up projections \citep{houlsby2019parameter}
\begin{equation}
\mathrm{Adapter}_{l_j}(u)
=W_{2,l_j}\mathrm{SiLU}(W_{1,l_j}u)
\end{equation}
where $W_{1,l_j}\in\mathbb{R}^{d_r\times d_{\mathrm{model}}}$ and
$W_{2,l_j}\in\mathbb{R}^{d_{\mathrm{model}}\times d_r}$ are the down
and up projections, respectively, with $d_r<d_{\mathrm{model}}$.
Given the execution gate $g_{l_j}$ defined below, the routed block
output is
\begin{equation}
h_{l_j}=
\begin{cases}
\bar{h}_{l_j}
+g_{l_j}\mathrm{FFN}_{l_j}(u_{l_j}),
& g_{l_j}\geq\tau_{\mathrm{exec}}\\
\bar{h}_{l_j}
+(1-g_{l_j})\mathrm{Adapter}_{l_j}(u_{l_j}),
& g_{l_j}<\tau_{\mathrm{exec}}
\end{cases}
\label{eq:ffn_route}
\end{equation}
Here, $g_{l_j}\in(0,1)$ and $\tau_{\mathrm{exec}}$ is the common
execution threshold. Training and inference use the same thresholded
single branch computation. Although the threshold operation is not
differentiated, scaling the selected branch by $g_{l_j}$ or
$1-g_{l_j}$ carries gradients to the router.

\subsection{HeRo Router}

\paragraph{Router state.}
At routed block $l_j$, HeRo combines an encoding of the current
representation with a path vector comprising six scalar features
\begin{equation}
x_{l_j}^{\mathrm{path}}
=\left[r_{l_j},q_{l_j}^{\mathrm{rem}},d_{l_j},\gamma_{l_j},
p_{l_{j-1}}^{\mathrm{hist}},m_{l_{j-1}}^{\mathrm{hist}}\right]^\top
\in\mathbb{R}^{6}
\label{eq:path_vector}
\end{equation}
The first two entries locate the current decision along routed depth,
the next two describe the latest completed residual stream transition,
and the final two retain routing scores from preceding blocks. The
current representation is encoded as
\begin{equation}
z_{l_j}^{h}
=W_{hu}\mathrm{SiLU}\!\left(
W_{hd}\mathrm{RMSNorm}(\bar{h}_{l_j})\right)
\in\mathbb{R}^{d_h}
\end{equation}
The two depth features specify the normalized routing progress and the
fraction of routed blocks remaining after the current decision
\begin{equation}
r_{l_j}=\frac{j}{L_R},
\qquad
q_{l_j}^{\mathrm{rem}}=\frac{L_R-j}{L_R}
\end{equation}
For $j\geq2$, the latest completed residual stream transition between
consecutive routed decision points and its magnitude are
\begin{equation}
\delta_{l_j}
=\bar{h}_{l_j}-\bar{h}_{l_{j-1}},
\quad
\delta_{l_1}=0
\qquad
d_{l_j}=\|\delta_{l_j}\|_2
\end{equation}
and its cosine similarity with the preceding transition is
\begin{equation}
\gamma_{l_j}=
\begin{cases}
0, & j\leq2\\
\displaystyle
\frac{\langle\delta_{l_j},\delta_{l_{j-1}}\rangle}
{\|\delta_{l_j}\|_2\|\delta_{l_{j-1}}\|_2+\varepsilon},
& j>2
\end{cases}
\label{eq:transition_features}
\end{equation}
The previous history score $p_{l_{j-1}}^{\mathrm{hist}}$ and cumulative
history score $m_{l_{j-1}}^{\mathrm{hist}}$ are defined by the gate
recurrence below, with
$p_{l_0}^{\mathrm{hist}}=m_{l_0}^{\mathrm{hist}}=1$. The path encoder
embeds the six scalar features, and its output is concatenated with the
current representation encoding to form the router state
\begin{equation}
\begin{aligned}
z_{l_j}^{a}
&=\mathrm{FFN}_{\mathrm{path}}(x_{l_j}^{\mathrm{path}})\\
s_{l_j}
&=[z_{l_j}^{h};z_{l_j}^{a}]
\in\mathbb{R}^{d_s}
\end{aligned}
\label{eq:router_state}
\end{equation}

\paragraph{Router memory mechanism.}
We implement the router memory mechanism with kernelized linear
attention \citep{katharopoulos2020transformers} over model depth. At each
routed block, it reads accumulated states before writing $s_{l_j}$, so
the context $c_{l_j}$ contains information only from preceding routed
blocks. The query, key, and value projections are
\begin{equation}
q_{l_j}=W_qs_{l_j},\qquad
k_{l_j}=W_ks_{l_j},\qquad
v_{l_j}=W_vs_{l_j}
\end{equation}
where $q_{l_j},k_{l_j},v_{l_j}\in\mathbb{R}^{d_m}$. Using the
positive feature map $\phi(x)=\mathrm{ELU}(x)+1$, let
$S_{l_{j-1}}$ and $\zeta_{l_{j-1}}$ denote the memory matrix and
normalizer accumulated before block $l_j$. The memory context is
\begin{equation}
c_{l_j}
=\frac{\phi(q_{l_j})^\top S_{l_{j-1}}}
{\phi(q_{l_j})^\top\zeta_{l_{j-1}}+\varepsilon}
\label{eq:memory_read}
\end{equation}
The current router state is written with the block specific forgetting
factor $\rho_{l_j}=\sigma(\eta_{l_j})$
\begin{equation}
\begin{aligned}
S_{l_j}
&=\rho_{l_j}S_{l_{j-1}}
+\phi(k_{l_j})v_{l_j}^{\top}\\
\zeta_{l_j}
&=\rho_{l_j}\zeta_{l_{j-1}}+\phi(k_{l_j})
\end{aligned}
\label{eq:memory_update}
\end{equation}
The read-before-write order gives the memory the following property: the context available to the current gate contains only states produced by earlier routed blocks.
For each token, $S_{l_0}=\mathbf{0}_{d_m\times d_m}$ and
$\zeta_{l_0}=\mathbf{0}_{d_m}$ are initialized at the start of a
forward pass and updated only across the routed blocks. All encoder,
memory, and router head parameters are shared across routed blocks,
whereas the adapters and forgetting factors are block specific.

\paragraph{Gate computation.}
The history head compares the current router state with the retrieved
context. Because no preceding context is available at the first routed
block, we set $\nu_{l_1}=0$. For $j>1$, the cosine dissimilarity between
the current state and retrieved context is
\begin{equation}
\nu_{l_j}=1-\cos(W_ns_{l_j},W_cc_{l_j})
\end{equation}
where $W_n$ and $W_c$ project the state and context into a common
comparison space. The state, context, and dissimilarity are then mapped
to the history logit $a_{l_j}$ and the recurrent history score
\begin{equation}
a_{l_j}
=f_{\theta}([s_{l_j};c_{l_j};\nu_{l_j}]),
\qquad
p_{l_j}^{\mathrm{hist}}=\sigma(a_{l_j}/T_{\mathrm{hist}})
\end{equation}
The cumulative product retains these scores for subsequent routed
blocks
\begin{equation}
m_{l_j}^{\mathrm{hist}}
=m_{l_{j-1}}^{\mathrm{hist}}p_{l_j}^{\mathrm{hist}}
\end{equation}
The local head operates directly on the representation after attention.
Combining its logit with the history logit gives the execution gate
\begin{equation}
b_{l_j}=f_{\psi}(\bar{h}_{l_j}),
\qquad
g_{l_j}=\sigma(a_{l_j}+b_{l_j})
\label{eq:final_gate}
\end{equation}

\subsection{Training Objective}

The backbone is frozen, while the adapters, state encoders, depth
memory, and router heads are optimized using the language modeling loss
and a gate utilization penalty
\begin{equation}
\mathcal{L}
=\mathcal{L}_{\mathrm{LM}}+\alpha_t\mathcal{L}_{\mathrm{skip}}
\end{equation}
For a batch of size $B$ and sequence length $T$, with $T-1$
next token prediction positions per sequence, the utilization penalty
is
\begin{equation}
\mathcal{L}_{\mathrm{skip}}
 =\frac{1}{B(T-1)}
\sum_{b=1}^{B}\sum_{t=1}^{T-1}
\left(\sum_{j=1}^{L_R}g_{b,l_j,t}\right)^2
\end{equation}
The inner sum couples gate values across routed blocks, and the square
penalizes large total FFN use for each token. The coefficient $\alpha_t$
sets its weight relative to the language modeling loss.

\begin{table*}[!t]
\centering
\small
\setlength{\tabcolsep}{2.4pt}
\renewcommand{\arraystretch}{1.08}
\begin{tabularx}{\textwidth}{@{}>{\raggedright\arraybackslash}p{0.18\textwidth}>{\raggedright\arraybackslash}p{0.12\textwidth}CCCCCCCC@{}}
\toprule
Method & Param skip & Retain & \shortstack{OBQA\\acc\_norm} & \shortstack{PIQA\\acc} & \shortstack{BoolQ\\acc} & \shortstack{ARC-E\\acc} & \shortstack{ARC-C\\acc} & \shortstack{WinoG\\acc} & \shortstack{HellaS\\acc\_norm} \\
\midrule
Llama-3.1-8B / Dense & 0\% & 100.00\% & 45.20 & 79.82 & 83.79 & 84.74 & 67.56 & 75.37 & 78.26 \\
Llama-3.1-8B / LoRA  & 0\% & 99.87\% & 43.80 & 77.49 & 79.76 & 84.55 & 77.48 & 74.69 & 75.42 \\
ShortGPT & 25.0\% & 54.87\% & 28.00 & 58.76 & 37.77 & 38.05 & 31.40 & 54.14 & 31.50 \\
Shortened-Taylor & 25.0\% & 54.93\% & 28.20 & 58.87 & 37.77 & 38.05 & 31.31 & 54.06 & 31.53 \\
SliceGPT & 24.6\% & 55.82\% & 30.40 & 57.83 & 37.83 & 38.64 & 25.85 & 55.17 & 38.19 \\
D-LLM & 25.0\% & 57.44\% & 30.20 & 57.40 & 50.36 & 37.12 & 28.16 & 52.49 & 37.64 \\
SkipGPT-Joint & 25.3\% & 60.91\% & 31.50 & 60.13 & 50.74 & 37.21 & 28.66 & 52.34 & 50.87 \\
MoD-D & 25.0\% & 61.52\% & 31.60 & 64.25 & 50.28 & 37.67 & 28.24 & 52.41 & 50.44 \\
Shortened-PPL & 25.0\% & 67.72\% & 33.60 & 71.87 & 42.08 & 57.07 & 31.74 & 53.51 & 57.98 \\
LLM-Pruner & 24.5\% & 70.38\% & 37.20 & 72.03 & 56.79 & 51.22 & 31.31 & 56.99 & 54.75 \\
LaCo & 24.5\% & 72.38\% & 31.20 & 66.16 & 71.13 & 48.65 & 37.03 & 65.11 & 55.77 \\
SkipGPT-RT & 25.5\% & \underline{94.14\%} & \underline{44.20} & \underline{78.07} & \underline{74.06} & \underline{82.44} & \underline{53.41} & \underline{75.69} & \underline{76.87} \\
\rowcolor{highlighter}
\textbf{HeRo, ($\alpha_t=10^{-3}$)} & 26.87{\fontsize{6}{7.2}\selectfont$\pm$0.80\%} & \textbf{100.24{\fontsize{6}{7.2}\selectfont$\pm$0.59\%}} & \textbf{44.93{\fontsize{6}{7.2}\selectfont$\pm$0.42}} & \textbf{79.38{\fontsize{6}{7.2}\selectfont$\pm$0.54}} & \textbf{82.63{\fontsize{6}{7.2}\selectfont$\pm$0.95}} & \textbf{84.41{\fontsize{6}{7.2}\selectfont$\pm$1.64}} & \textbf{70.56{\fontsize{6}{7.2}\selectfont$\pm$0.34}} & \textbf{76.69{\fontsize{6}{7.2}\selectfont$\pm$0.59}} & \textbf{76.98{\fontsize{6}{7.2}\selectfont$\pm$0.10}} \\
\midrule
Llama-2-7B / Dense & 0\% & 100.00\% & 44.20 & 78.07 & 71.62 & 81.36 & 52.47 & 74.19 & 78.93 \\
Llama-2-7B / LoRA  & 0\% & 100.09\% & 44.00 & 76.62 & 77.61 & 80.27 & 52.12 & 72.94 & 77.55 \\
Shortened-PPL & 25.00\% & 73.76\% & 33.60 & 70.40 & 61.07 & 55.05 & 29.44 & 52.88 & 55.12 \\
MoD-D & 25.00\% & 79.17\% & 28.00 & 69.86 & 62.29 & \underline{72.81} & 43.34 & 50.28 & 58.80 \\
LLM-Pruner & 25.30\% & 79.61\% & 39.00 & \underline{73.45} & 54.71 & 58.63 & 37.03 & 58.56 & 60.77 \\
LaCo & 25.00\% & 83.74\% & 36.60 & 65.72 & \textbf{74.37} & 58.92 & 38.14 & \underline{67.32} & 62.76 \\
SkipGPT-RT & 25.50\% & \underline{90.33\%} & \underline{39.60} & 72.20 & \underline{68.81} & \textbf{76.52} & \underline{44.37} & 63.54 & \textbf{70.96} \\
\rowcolor{highlighter}
\textbf{HeRo, ($\alpha_t=10^{-3}$)} & 27.87{\fontsize{6}{7.2}\selectfont$\pm$0.82\%} & \textbf{94.38{\fontsize{6}{7.2}\selectfont$\pm$0.54\%}} & \textbf{41.67{\fontsize{6}{7.2}\selectfont$\pm$0.36}} & \textbf{76.51{\fontsize{6}{7.2}\selectfont$\pm$0.49}} & 67.59{\fontsize{6}{7.2}\selectfont$\pm$0.94} & 67.79{\fontsize{6}{7.2}\selectfont$\pm$1.33} & \textbf{53.92{\fontsize{6}{7.2}\selectfont$\pm$0.40}} & \textbf{73.23{\fontsize{6}{7.2}\selectfont$\pm$0.61}} & \underline{70.60{\fontsize{6}{7.2}\selectfont$\pm$0.11}} \\
\midrule
Llama-2-13B / Dense & 0\% & 100.00\% & 45.20 & 79.11 & 80.52 & 84.68 & 59.47 & 76.16 & 82.23 \\
Llama-2-13B / LoRA  & 0\% & 100.05\% & 45.20 & 79.11 & 80.06 & 84.97 & 59.72 & 76.24 & 82.26 \\
LaCo & 25.00\% & 75.05\% & 38.80 & 72.74 & 44.46 & 62.84 & 35.07 & 62.75 & 63.11 \\
MoD-D & 25.00\% & 81.18\% & 37.20 & 71.49 & 54.13 & \textbf{78.24} & \underline{50.34} & 67.25 & 51.83 \\
Shortened-PPL & 25.00\% & 82.95\% & 39.40 & 73.12 & 62.57 & 69.19 & 41.13 & 67.17 & 69.31 \\
LLM-Pruner & 24.90\% & 86.92\% & 44.00 & \underline{76.99} & 59.88 & 72.60 & 45.82 & 65.82 & 74.15 \\
SkipGPT-RT & 25.40\% & \underline{92.43\%} & \textbf{46.00} & 76.88 & \textbf{74.37} & \underline{77.69} & 47.08 & \underline{71.90} & \underline{74.33} \\
\rowcolor{highlighter}
\textbf{HeRo, ($\alpha_t=10^{-3}$)} & 28.38{\fontsize{6}{7.2}\selectfont$\pm$0.76\%} & \textbf{94.49{\fontsize{6}{7.2}\selectfont$\pm$0.59\%}} & \underline{45.67{\fontsize{6}{7.2}\selectfont$\pm$0.40}} & \textbf{77.60{\fontsize{6}{7.2}\selectfont$\pm$0.55}} & \underline{73.83{\fontsize{6}{7.2}\selectfont$\pm$1.11}} & 75.25{\fontsize{6}{7.2}\selectfont$\pm$1.34} & \textbf{52.24{\fontsize{6}{7.2}\selectfont$\pm$0.38}} & \textbf{76.23{\fontsize{6}{7.2}\selectfont$\pm$0.49}} & \textbf{77.32{\fontsize{6}{7.2}\selectfont$\pm$0.10}} \\
\midrule
Llama-3.1-8B / Dense & 0\% & 100.00\% & 45.20 & 79.82 & 83.79 & 84.74 & 67.56 & 75.37 & 78.26 \\
LaCo & 40.70\% & 54.28\% & 27.60 & 56.31 & 55.11 & 30.22 & 25.85 & 51.54 & 31.53 \\
LLM-Pruner & 39.90\% & 56.83\% & 29.20 & 64.09 & 50.52 & 36.87 & 23.46 & 51.30 & 36.23 \\
Shortened-PPL & 40.60\% & 59.11\% & 27.00 & 61.48 & 57.13 & 39.69 & 26.45 & 52.41 & 41.74 \\
MoD-D & 40.00\% & 63.14\% & 33.00 & 65.56 & 50.28 & 38.09 & 30.20 & 51.38 & 54.01 \\
SkipGPT-RT & 40.20\% & \underline{81.15\%} & \underline{38.00} & \underline{73.34} & \underline{60.37} & \textbf{77.53} & \underline{45.65} & \underline{59.35} & \underline{64.36} \\
\rowcolor{highlighter}
\textbf{HeRo, ($\alpha_t=10^{-3}$)} & 38.82{\fontsize{6}{7.2}\selectfont$\pm$0.84\%} & \textbf{97.01{\fontsize{6}{7.2}\selectfont$\pm$0.70\%}} & \textbf{42.47{\fontsize{6}{7.2}\selectfont$\pm$0.44}} & \textbf{79.45{\fontsize{6}{7.2}\selectfont$\pm$0.54}} & \textbf{82.06{\fontsize{6}{7.2}\selectfont$\pm$0.98}} & \underline{76.56{\fontsize{6}{7.2}\selectfont$\pm$1.38}} & \textbf{69.30{\fontsize{6}{7.2}\selectfont$\pm$0.30}} & \textbf{74.10{\fontsize{6}{7.2}\selectfont$\pm$0.70}} & \textbf{75.60{\fontsize{6}{7.2}\selectfont$\pm$0.09}} \\
\bottomrule
\end{tabularx}
\caption{Comparison on seven benchmarks at target parameter skipping budgets of 25\% and 40\%. Results are grouped by backbone and target budget. HeRo learns an execution path for each token over the candidate MLP modules associated with each target budget. The first group includes all compared methods. Each remaining group reports Dense and the six methods with the highest Retain in the first group. Retain is the mean task score relative to the corresponding Dense score, expressed as a percentage. Bold and underline indicate the highest and second highest scores among methods that skip parameters within each group. HeRo results are means with standard deviations over random seeds 42, 43, and 44. Dense, LoRA $r{=}128$ denotes the dense backbone fine-tuned with the same adapter and Tulu-3 SFT training without layer skipping, serving as a LoRA reference.}
\label{tab:main_results}
\end{table*}

\section{Experiments}

We compare HeRo with static pruning and dynamic routing baselines under matched target parameter skipping budgets. We then vary the skipping coefficient $\alpha_t$ to characterize its effect on task performance and parameter skipping, remove individual path state components to assess the roles of path features, memory read, and history conditioning, and examine sensitivity to the memory dimension. Together, these experiments evaluate whether explicit path conditioning improves task performance at comparable realized parameter skipping rates.

\begin{table*}[!t]
\centering
\small
\setlength{\tabcolsep}{2.4pt}
\renewcommand{\arraystretch}{1.08}
\begin{tabularx}{\textwidth}{@{}>{\raggedright\arraybackslash}p{0.18\textwidth}>{\raggedright\arraybackslash}p{0.12\textwidth}CCCCCCCC@{}}
\toprule
Method & Param skip & Retain & \shortstack{OBQA\\acc\_norm} & \shortstack{PIQA\\acc} & \shortstack{BoolQ\\acc} & \shortstack{ARC-E\\acc} & \shortstack{ARC-C\\acc} & \shortstack{WinoG\\acc} & \shortstack{HellaS\\acc\_norm} \\
\midrule
$\alpha_t=1\mathrm{e}{-4}$ & 13.19\% & \underline{101.66\%} & 46.20 & \underline{80.14} & \textbf{82.75} & \textbf{86.49} & \textbf{73.58} & \underline{77.03} & \underline{77.36} \\
$\alpha_t=3\mathrm{e}{-4}$ & 18.14\% & \textbf{101.73\%} & \underline{46.80} & \textbf{80.36} & \underline{82.66} & \underline{85.44} & \underline{73.24} & \textbf{77.51} & \textbf{77.44} \\
$\alpha_t=5\mathrm{e}{-4}$ & 23.40\% & 101.28\% & \textbf{47.00} & 79.76 & 81.38 & 85.09 & \underline{73.24} & 76.95 & 77.32 \\
\rowcolor{highlighter}
$\alpha_t=1\mathrm{e}{-3}$ & 26.87\% & 100.00\% & 44.93 & 79.38 & 82.63 & 84.41 & 70.56 & 76.69 & 76.98 \\
\bottomrule
\end{tabularx}
\caption{Effect of $\alpha_t$ on parameter skipping and task accuracy for Meta-Llama-3.1-8B-Instruct. Retain is the mean of the seven task score ratios relative to the $\alpha_t=10^{-3}$ configuration within this sweep, expressed as a percentage. Results are means over random seeds 42, 43, and 44. Bold and underline indicate the highest and second highest values.}
\label{tab:alpha_sweep}
\end{table*}

\subsection{Experimental Setup}

\noindent\textbf{Models and routing configuration.}
Meta-Llama-3.1-8B-Instruct \citep{grattafiori2024llama3} is the primary backbone for the controlled analyses. Table~\ref{tab:main_results} also reports results for Llama-2-7B and Llama-2-13B. Unless stated otherwise, the configuration below refers to Meta-Llama-3.1-8B-Instruct. For the 25\% target-budget groups in Table~\ref{tab:main_results}, the candidate MLP modules are the latter half of each backbone, while earlier layers and all attention modules remain dense. For the Llama-3.1-8B 40\% target-budget group, the candidate set is expanded to include layers 9 through 32. This placement follows prior evidence that earlier layers are more sensitive to removal \citep{men-etal-2025-shortgpt}; the candidate set defines routing eligibility, while HeRo learns the execution path for each token. At each routed layer, HeRo selects either the pretrained MLP or a bottleneck adapter with intermediate width $d_r=896$ from the representation after attention. The router state has 64 dimensions, comprising 48 dimensions for the hidden representation and 16 dimensions for auxiliary path features. The history and local routing heads each use a hidden width of 256. The depth memory has one head with $q,k,v\in\mathbb{R}^{64}$; for each token, it maintains a $64\times64$ memory matrix and a normalizer of dimension 64. Each routed layer has a learned forgetting parameter $\eta_{l_j}$, initialized such that $\rho_{l_j}=\sigma(\eta_{l_j})\approx0.98$. The principal configuration uses $\alpha_t=10^{-3}$ and a target parameter skipping budget of 25\%.

\noindent\textbf{Training configuration.}
The backbone remains frozen. We update only the router state encoder, depth memory, history and local routing heads, and adapters. Training uses all 939,344 examples in the Tulu-3 SFT mixture \citep{ivison2025tulu} for one epoch, corresponding to approximately 7,339 optimizer updates at a global batch size of 128. We apply the chat template associated with each backbone, truncate sequences to 2048 tokens, and compute the language modeling loss over all tokens. Optimization uses AdamW with a learning rate of $10^{-4}$, weight decay 0.01, $\beta_1=0.9$, $\beta_2=0.999$, $\epsilon=10^{-8}$, linear warmup over the first 3\% of updates, and gradient clipping at 1.0. We use gradient checkpointing and bfloat16 mixed precision across four NVIDIA RTX 4090D GPUs with 48\,GB of memory each. Each training run takes approximately 12 hours. We set the execution threshold to $\tau_{\mathrm{exec}}=0.5$ and the history temperature to $T_{\mathrm{hist}}=0.9$. For every setting in the main comparison, we train HeRo with random seeds 42, 43, and 44 and report the mean and standard deviation. Training and evaluation both use the thresholded branch decision in Equation~\ref{eq:ffn_route}; the gate scaling supplies gradients to the router during training.

\noindent\textbf{Benchmarks and metrics.}
We use the lm-evaluation-harness framework \citep{gao2024framework}. The main comparison covers seven benchmarks: OpenBookQA \citep{mihaylov2018can} and ARC-Easy and ARC-Challenge \citep{clark2018think} for science question answering; PIQA \citep{bisk2020piqa} for physical commonsense; BoolQ \citep{clark2019boolq} for question answering over passages; WinoGrande \citep{sakaguchi2021winogrande} for commonsense coreference; and HellaSwag \citep{zellers2019hellaswag} for plausible continuation selection. We report normalized accuracy for OpenBookQA and HellaSwag and accuracy for the other five tasks. The ablation suite additionally includes GSM8K \citep{cobbe2021gsm8k}, evaluated by exact match after chain of thought generation, and HumanEval \citep{chen2021humaneval}, evaluated by pass@1 under greedy decoding.

\begin{table*}[!t]
\centering
\small
\setlength{\tabcolsep}{0.8pt}
\renewcommand{\arraystretch}{1.08}
\begin{tabularx}{\textwidth}{@{}>{\raggedright\arraybackslash}p{0.12\textwidth}CCCCCCCCCCC@{}}
\toprule
Method & Param skip & Retain & \shortstack{GSM8K\\exact\_match} & \shortstack{ARC-C\\acc} & \shortstack{OBQA\\acc\_norm} & \shortstack{HumE\\pass@1} & \shortstack{BoolQ\\acc} & \shortstack{WinoG.\\acc} & \shortstack{ARC-E\\acc} & \shortstack{PIQA\\acc} & \shortstack{HellaS\\acc\_norm} \\
\midrule
 w/o Pos State & 26.35\% & 97.57\% & \underline{80.52} & 68.23 & 42.20 & 61.59 & 79.02 & 76.09 & 83.68 & \underline{79.82} & \underline{77.35} \\
 w/o Aux State & 27.02\% & 97.32\% & 79.38 & 68.56 & 42.20 & 60.37 & \underline{79.42} & 75.77 & \underline{84.56} & 79.43 & 77.32 \\
 w/o History & 26.41\% & 97.35\% & 79.38 & 67.89 & \underline{42.80} & 59.76 & 78.41 & \underline{76.64} & \textbf{84.74} & \textbf{79.92} & \textbf{77.38} \\
 w/o MemRead & 26.28\% & \underline{98.02\%} & 80.14 & \underline{69.23} & 42.60 & \underline{62.20} & 79.20 & 76.48 & 84.39 & 79.60 & 77.33 \\
\rowcolor{highlighter}
\textbf{HeRo} & 26.87\% & \textbf{100.00\%} & \textbf{82.11} & \textbf{70.56} & \textbf{44.93} & \textbf{65.24} & \textbf{82.63} & \textbf{76.69} & 84.41 & 79.38 & 76.98 \\
\bottomrule
\end{tabularx}
\caption{Component ablations of HeRo on Meta-Llama-3.1-8B-Instruct using the extended benchmark suite. Param skip is the realized parameter skipping rate, averaged over tokens and normalized by the total parameter count. Retain is the mean of the nine task score ratios relative to the full HeRo configuration, expressed as a percentage.}
\label{tab:core_ablation}
\end{table*}

\noindent\textbf{Evaluation protocol.}
GSM8K uses eight chain of thought demonstrations and a maximum generation length of 1024 tokens. ARC-Easy and ARC-Challenge use 25 demonstrations, while HellaSwag and WinoGrande use five. OpenBookQA, PIQA, BoolQ, and HumanEval are evaluated without demonstrations. Chat templates are enabled for tasks that use instruction formatting, and demonstrations are rendered as multiturn conversations when required by the task definition.

\noindent\textbf{Baselines and comparison budgets.}
The static pruning baselines are ShortGPT \citep{men-etal-2025-shortgpt}, the PPL and Taylor variants of Shortened LLaMA \citep{kim2024shortened}, LaCo \citep{yang-etal-2024-laco}, LLM-Pruner \citep{ma2023llmpruner}, and SliceGPT \citep{ashkboos2024slicegpt}. Dynamic baselines include MoD-D, a variant of Mixture of Depths introduced by SkipGPT that routes attention and MLP modules separately \citep{raposo2024mixture,zhao2025skipgpt}; D-LLM \citep{jiang2024dllm}; and the joint training and router tuning variants of SkipGPT \citep{zhao2025skipgpt}. These methods bypass different computational units, including complete Transformer layers, attention modules, and MLP modules. We therefore group comparisons by target parameter skipping budget and report the realized parameter skipping rate for every method. This rate is the proportion of model parameters that do not participate in computation, averaged over tokens and normalized by the total parameter count.

\subsection{Main Results}

\noindent\textbf{Performance at matched parameter budgets.}
Table~\ref{tab:main_results} shows that HeRo achieves the highest aggregate retention among the evaluated parameter skipping methods across all three backbones. On Llama-3.1-8B at the 25\% target budget, HeRo matches the dense model's aggregate performance at a realized parameter skipping rate of 26.87\% and leads all competing pruning and routing methods on every benchmark. HeRo also ranks first in Retain on Llama-2-7B and Llama-2-13B. The repeated advantage across model sizes shows that its performance is not specific to a single backbone.

\noindent\textbf{Behavior under a tighter computation budget.}
The 40\% target budget provides a more demanding test because more residual updates are bypassed before subsequent routing decisions are made. HeRo retains 97.01\% of dense performance in this setting, compared with 81.15\% for SkipGPT-RT at a similar realized parameter skipping rate, and leads on six of the seven benchmarks. The margin over the strongest dynamic baseline therefore grows substantially as the target budget becomes more restrictive. This pattern is consistent with the motivation for HeRo: as more MLP updates are bypassed, subsequent routers receive representations shaped by increasingly diverse execution paths, making an explicit summary of earlier routing decisions more informative for downstream allocation.

\subsection{Ablation Studies}

\noindent\textbf{Effect of the skipping coefficient.}
Table~\ref{tab:alpha_sweep} shows that $\alpha_t$ provides direct control over the operating point of HeRo. Increasing the coefficient raises the realized parameter skipping rate monotonically from 13.19\% to 26.87\%, while Retain varies by only 1.73 points across the sweep. Lower coefficients yield slightly higher Retain, whereas $\alpha_t=10^{-3}$ reaches the target budget with only a modest reduction in aggregate performance. We therefore use this setting in the main comparison.

\noindent\textbf{Path state components.}
Table~\ref{tab:core_ablation} isolates the information flow that distinguishes HeRo from a local router. The \textit{w/o History} variant removes the history branch and bases each decision on the local logit alone. The \textit{w/o MemRead} variant retains the history head but removes its access to accumulated memory. The remaining variants remove either the auxiliary path encoder or the routing progress and residual transition features supplied to it. All other training and routing settings are fixed.

The complete model achieves the highest Retain at a parameter skipping rate close to the maximum in the table. Every ablation lowers aggregate performance. Removing the auxiliary path encoder slightly increases Param skip, while the other ablations reduce both Retain and Param skip. The quality advantage of the complete model is therefore not explained by a more conservative routing policy. These results show that state construction and history conditioning both contribute to routing quality.

The memory read provides a second, complementary source of evidence. Retaining the history head without access to the accumulated memory still reduces both Retain and Param skip, so the history head alone does not recover the full model's routing quality. Taken together, the ablations support the complete HeRo design as a sequence of complementary operations: path features describe the current routing context, the memory aggregates that context across depth, and the history head uses the accumulated state to coordinate the next branch decision.

\begin{table}[!ht]
\centering
\small
\setlength{\tabcolsep}{4.2pt}
\renewcommand{\arraystretch}{1.08}
\begin{tabular}{@{}lccc@{}}
\toprule
Memory dim. & Param skip & Retain & Avg. acc \\
\midrule
32  & 26.36\% & \underline{100.08\%} & 73.61 \\
64  & 26.87\% & 100.00\% & \underline{73.65} \\
128 & 26.29\% & 100.03\% & 73.62 \\
256 & 26.54\% & \textbf{100.12\%} & \textbf{73.66} \\
\bottomrule
\end{tabular}
\caption{Effect of memory dimension on Meta-Llama-3.1-8B-Instruct. All other settings follow the main configuration.}
\label{tab:hero_llama31_memory_ablation}
\end{table}

\noindent\textbf{Memory dimension.}
Table~\ref{tab:hero_llama31_memory_ablation} shows that no memory dimension dominates all three aggregate measures. A dimension of 256 yields the highest Retain and average accuracy with the second highest parameter skipping rate, whereas a dimension of 64 yields the highest parameter skipping rate but the lowest Retain. Across the tested range, Retain and average accuracy vary by only 0.12 and 0.05 points, indicating limited sensitivity to the memory dimension.

\section{Conclusion}

 We introduced History-Aware Routing (HeRo), a dynamic routing framework that incorporates a \textbf{router memory mechanism} based on \textbf{linear attention} over model depth. The mechanism aggregates preceding routing scores and their induced residual updates into an explicit routing state, allowing each router to condition its decision on both the current hidden state and the routing information accumulated across preceding layers. Across Llama 3.1-8B, Llama 2-7B, and Llama 2-13B, HeRo consistently outperformed all 10 baseline methods while training only lightweight routers and adapters and keeping the \textbf{backbone frozen}, demonstrating consistent effectiveness across model families and scales. On Llama 3.1-8B, \textbf{HeRo bypassed 26.87\% of model parameters during inference while achieving 100.24\% of dense model performance} across seven benchmarks; under a more restrictive computation budget, \textbf{it retained 97.01\% of dense performance while bypassing 38.82\% of parameters}. Ablations further showed that removing the routing history head led to performance losses across six benchmarks, with particularly notable losses on HumanEval and GSM8K. These results confirm that \textbf{explicitly maintaining routing information} through the router memory mechanism improves the coordination of routing decisions across model depth, enabling more accurate and selective dynamic routing than routing based solely on the current hidden state.

\bibliography{aaai2027}

\end{document}